\documentclass[11pt]{tibop-article}
\TIBauthor[Mishra et al.]{Shivam Mishra\affOne\orcidlink{0009-0005-1245-6501}\and Dhannu Ram Meena\affOne\orcidlink{0009-0003-3953-4675}\and Muneendra Ojha\affOne\orcidlink{0000-0002-3913-1185}\and Krishna Pratap Singh\affOne\orcidlink{0000-0003-3347-8521}\lastand \\ Kuldeep Singh\affOne\orcidlink{0009-0009-9781-0862} }
\TIBaffiliations{\affOne AIMS Lab, Indian Institute of Information Technology Allahabad, Prayagraj, UP, India\medskip\\
	*Correspondence: Shivam Mishra, m.shivam0704@gmail.com
}
\usepackage{amsmath}
\usepackage{tikz}
\usepackage{float}
\usetikzlibrary{positioning,arrows.meta,shapes.geometric}
\usepackage{tabularx}
\usepackage{array}
\newcolumntype{C}{>{\centering\arraybackslash}X}

\usepackage{algorithm}
\usepackage{algpseudocode}
\usepackage{tikz}
\usetikzlibrary{positioning,fit,arrows.meta}
\usepackage{biblatex}
\usepackage{graphicx}

\TIBtitle{pro-team at LLMs4OL 2026 Tasks Flagship and Reuse: Retrieval-Augmented Generation and Vocabulary-Constrained Filtering for Ontology Learning}
\TIBbundlename[Open Conf Proc X (2026) "LLMs4OL 2026: The 3rd Large Language Models for Ontology Learning Challenge at the 25th ISWC"]{LLMs4OL 2026: The 3rd Large Language Models for Ontology Learning Challenge at the 25th ISWC}
\TIBconferencesession{LLMs4OL 2026 Task Participant Papers}

\TIBabstract{Ontology learning from text remains challenging despite significant progress in Large Language Models (LLMs), which can hallucinate domain terms, produce inconsistent formats, and favor hierarchical over associative relations. In the LLMs4OL 2026 Challenge, we address both the End-to-End Flagship Task (Task A) and Ontology Extension Reuse Task (Task B) using an offline retrieval-augmented few-shot prompting pipeline. Our system employs Qwen2.5-14B-Instruct with all-MiniLM-L6-v2 for demonstration retrieval, selecting the top-5 examples for Task A and top-2 for Task B. A left-truncated context-windowing strategy preserves task instructions within long prompts. For Task B, generated triples undergo deterministic vocabulary-constrained filtering, retaining triples when at least one endpoint belongs to the sample’s closed term/type vocabulary and removing duplicates of the initial ontology. The approach achieves Semantic Graph Similarity of 0.8692, Term-Typing F1 of 0.9200, and Taxonomy Discovery F1 of 0.8540 on Task B, while Task A achieves 0.7416 Semantic Graph Similarity. However, no non-taxonomic relations are extracted, highlighting limitations of closed, taxonomy-oriented relation vocabularies.}
\TIBkeywords{Ontology Learning, Large Language Models, Retrieval-Augmented Generation (RAG), Vocabulary-Constrained Filtering, Knowledge Representation.}
\usepackage{xfrac}
\begin{document}
\maketitle

\section{Introduction}

Formal ontologies provide machine-readable representations of domain knowledge by defining concepts, semantic relationships, and logical constraints that enable knowledge integration, interoperability, and automated reasoning across heterogeneous information systems~\cite{li2026large}. They serve as the semantic backbone of numerous intelligent applications, including semantic search, recommendation systems, knowledge graphs, and digital libraries~\cite{aggarwal2026large}. However, constructing and maintaining ontologies remains a labor-intensive process that requires substantial domain expertise and continuous refinement as domain knowledge evolves~\cite{perera2024exploring}. With the rapid growth of scientific literature and unstructured textual resources, manual ontology engineering has become increasingly difficult to scale, motivating the development of automated Ontology Learning (OL) techniques~\cite{aggarwal2026large}. Ontology Learning aims to automatically or semi-automatically construct ontologies by extracting concepts, semantic relations, taxonomies, and other ontology components from textual data~\cite{perera2024exploring}. Traditional ontology learning methods relied on rule-based systems, statistical learning, and conventional natural language processing techniques for identifying ontology elements~\cite{perera2024exploring}. However, these methods achieved promising results in controlled environments; they often struggle with linguistic ambiguity, domain-specific terminology, and morphologically complex language structures, limiting their applicability across diverse domains ~\cite{terdalkar2019framework}, recent advancements in LLMs have significantly transformed ontology engineering by enabling zero-shot and few-shot reasoning over natural language without extensive task-specific feature engineering~\cite{li2026large}. Modern instruction-tuned LLMs can identify concepts, semantic relations, ontology mappings, and logical structures directly from textual documents, making them promising tools for automating ontology development~\cite{aggarwal2026large}. Nevertheless, recent surveys emphasize that effectively integrating LLMs into ontology engineering workflows remains an active research challenge because issues related to evaluation, reproducibility, and structured ontology construction have not yet been fully resolved~\cite{li2026large}.

To promote systematic evaluation of LLM-based ontology learning methods, the Large Language Models for Ontology Learning (LLMs4OL) paradigm introduced a standardized benchmark covering multiple ontology engineering tasks~\cite{10.1007/978-3-031-47240-4_22, BabaeiGiglou2024LLMs4OL}. The second edition of the challenge evaluated systems across four complementary tasks: Text-to-Ontology Extraction, Term Typing, Taxonomy Discovery, and Non-Taxonomic Relation Extraction~\cite{BabaeiGiglou2025LLMs4OL}. The third edition, which this paper addresses, restructures the benchmark around three integrated tracks that combine these primitives: the Flagship Task (Task A), requiring end-to-end construction of a primitive ontology from raw text; the Reuse Task (Task B), requiring incremental extension of a partial ontology; and the Taxonomy Task (Task C), requiring cross-domain taxonomy induction~\cite{BabaeiGiglou2026LLMs4OL}. Together, these tasks evaluate the complete ontology learning pipeline, from identifying ontology concepts in raw text to assigning semantic categories and discovering hierarchical and associative relationships among ontology entities. Among the ontology primitives these tasks target, accurate term extraction and semantic type assignment constitute the foundation of the ontology construction process, since they directly influence the quality of downstream taxonomy induction and relation extraction. Errors introduced during these early stages propagate to subsequent ontology learning tasks, reducing the overall consistency and usability of the generated ontology. Therefore, improving performance on the Flagship (Task A) and Reuse (Task B) tracks addressed in this paper remains essential for building reliable automated ontology learning systems.

Despite the remarkable reasoning capability of modern LLMs, several challenges continue to limit their effectiveness in ontology learning. Generative models frequently produce hallucinated or lexically inconsistent outputs that do not belong to the predefined ontology vocabulary, thereby reducing semantic consistency and increasing post-processing complexity~\cite{BabaeiGiglou2025LLMs4OL}. Existing studies also show that while LLMs perform well on hierarchical taxonomy prediction, they remain considerably less effective at identifying complex semantic and non-taxonomic relationships that require deeper contextual understanding~\cite{bakker2025ontology}. Furthermore, retrieval-based prompting often requires long contextual prompts containing multiple demonstrations, increasing the likelihood of context-window truncation and loss of important task instructions during inference~\cite{BabaeiGiglou2025LLMs4OL}.

To overcome the identified challenges,we proposes an offline Retrieval-Augmented Generation (RAG) framework for ontology learning that combines dense semantic retrieval with instruction-guided LLM inference. The retrieval module dynamically selects semantically similar training examples to provide high-quality contextual guidance during inference while maintaining a completely offline pipeline suitable for resource-constrained environments \cite{BabaeiGiglou2025LLMs4OL}. We further employ a left-truncated context management strategy to preserve task instructions when processing long prompts \cite{BabaeiGiglou2025LLMs4OL}. Finally, for tasks with closed-world assumptions, we introduce a deterministic vocabulary-constrained filtering step that strictly evaluates generated outputs against the predefined ontology vocabulary using exact lexical matching, dropping any invalid triples. Similar embedding-based semantic neighborhood approaches have demonstrated that constraining predictions to known semantic spaces significantly improves ontology classification performance \cite{yimmark2025t, lexical2025}.The primary contributions of this work are summarized as follows:

\begin{itemize}
    \item We propose an offline Retrieval-Augmented Generation framework for ontology learning that combines dense semantic retrieval with efficient prompt construction.

    \item We introduce a vocabulary-constrained filtering algorithm for the Reuse Task that restricts generated outputs to the canonical ontology vocabulary, eliminating out-of-vocabulary hallucinations by dropping invalid triples.
    \item We evaluate the proposed framework on the official LLMs4OL benchmark datasets for Task A (End-to-End Flagship) and Task B (Ontology Extension Reuse) using exact-match, fuzzy-match, and semantic similarity metrics.
    \item We provide a detailed error analysis to identify the remaining challenges in LLM-based ontology learning and discuss directions for future research.
\end{itemize}

The remainder of this paper is organized as follows. Section~2 reviews existing research on ontology learning and LLM-based ontology engineering. Section~3 presents the proposed methodology. Section~4 describes the experimental setup and reports the evaluation results. Section~5 discusses the findings and limitations of the proposed approach. Finally, Section~6 concludes the paper.

\section{Related Work}

Ontology Learning (OL) has evolved considerably over the past two decades, progressing from rule-based information extraction techniques to modern Large Language Model (LLM)-driven semantic reasoning. This section reviews the major developments in ontology learning, discusses recent advances in LLM-based ontology engineering, and positions the proposed Retrieval-Augmented Generation (RAG) framework within the existing literature.

\subsection{Classical Ontology Learning}

Traditional ontology learning focused on automatically extracting ontology elements such as concepts, taxonomies, semantic relations, and axioms from unstructured textual resources. Earlier approaches mainly used rule-based systems, statistical learning, and conventional natural language processing techniques, including tokenization, pos tagging, and NER, for ontology construction. These methods significantly reduced manual effort but depended heavily on handcrafted linguistic rules and domain-specific feature engineering, limiting their adaptability across domains~\cite{perera2024exploring}.

The challenges become even more pronounced when processing morphologically rich languages. Existing studies have shown that the limited availability of mature NLP tools, together with complex grammatical structures, substantially reduces the effectiveness of traditional ontology extraction systems and often requires explicit linguistic knowledge for accurate semantic interpretation~\cite{terdalkar2019framework}.

Although classical methods remain computationally efficient and interpretable, their limited contextual understanding prevents them from capturing implicit semantic relationships, motivating the adoption of deep learning and, more recently, large language models for ontology learning.

\subsection{Large Language Models for Ontology Engineering}

Recent advances in LLMs have substantially changed the ontology engineering landscape by enabling direct reasoning over natural language without manual feature engineering. Recent studies have shown that LLMs can effectively support ontology construction through zero-shot and few-shot reasoning while reducing the amount of manual annotation required~\cite{perera2024exploring}.

A comprehensive review of recent research further demonstrates that LLMs are now being applied across multiple stages of the ontology engineering lifecycle, including ontology generation, alignment, evaluation, maintenance, and documentation. The review also highlights the absence of standardized evaluation protocols and reproducible experimental workflows, indicating that LLM-based ontology engineering remains an active research area~\cite{li2026large}.

Experimental evaluations on scholarly ontology generation have further shown that several instruction-tuned open-source and proprietary LLMs achieve strong zero-shot performance for semantic relation identification while requiring relatively modest computational resources. These findings demonstrate the increasing capability of modern LLMs to support automated ontology construction across specialized knowledge domains~\cite{aggarwal2026large}.

\subsection{LLMs4OL Benchmark and Retrieval-Augmented Approaches}

To facilitate systematic evaluation of ontology learning systems, the LLMs4OL challenge introduced standardized benchmark datasets covering multiple ontology engineering tasks across several application domains~\cite{giglou2024llms4ol}. The second edition of the challenge evaluated systems on four complementary tasks, namely Text-to-Ontology Extraction, Term Typing, Taxonomy Discovery, and Non-Taxonomic Relation Extraction~\cite{beliaeva2025heterogeneous}, and the third edition consolidates these primitives into the integrated Flagship, Reuse, and Taxonomy tracks addressed in this work, thereby encouraging the development of more robust and scalable end-to-end ontology learning systems.

Recent retrieval-augmented approaches combine dense semantic retrieval with few-shot prompting to improve ontology learning. Instead of relying only on the internal knowledge of language models, these methods dynamically retrieve semantically similar training examples to guide inference, resulting in improved performance on term extraction and semantic typing tasks~\cite{BabaeiGiglou2025LLMs4OL}.Other combines semantic clustering with LLM inference by grouping semantically related concepts using transformer-based embeddings before prompt generation. This strategy reduces semantic ambiguity and improves prediction quality across multiple ontology learning tasks, highlighting the growing importance of retrieval and semantic representation learning in modern ontology learning pipelines~\cite{goyal2025silp_nlp}.

\subsection{Challenges in Ontology Learning}

While modern LLMs have impressive capabilities, several challenges remain that limit fully automated ontology construction. Recent research on ontology axiom identification has shown that LLMs perform significantly better on hierarchical subclass prediction compared to more expressive ontology axioms like domain, range, and disjointness constraints. These findings show that higher-level semantic reasoning is still much harder than taxonomy extraction~\cite{bakker2025ontology}.

Another important challenge is the lexical inconsistency introduced by generative models when predicting ontologies. Recently, we have shown that semantic embeddings based on transformers can be combined with classical k-Nearest Neighbors classification to improve the typing of ontology terms, where predictions are limited to a semantically meaningful neighborhood. Such embedding-based neighborhood classification always improves the prediction accuracy compared to the direct generative outputs.~\cite{yimmark2025t}.

\subsection{Research Gap}

The current literature shows promising advancements in applying LLMs on ontology learning, however, some limitations still exist. Classical ontology learning approaches lack a deep contextual understanding, while purely generative LLM-based systems are prone to generate hallucinated or lexically inconsistent ontology terms.~\cite{perera2024exploring}. Retrieval-augmented approaches improve contextual grounding but primarily focus on prompt engineering and semantic retrieval without explicitly enforcing ontology vocabulary consistency during generation~\cite{BabaeiGiglou2025LLMs4OL}. Similarly, embedding-based classification methods improve semantic prediction but are generally designed for individual tasks rather than an integrated ontology learning pipeline~\cite{yimmark2025t}.

Motivated by these limitations, our work combines dense Retrieval-Augmented Generation with a deterministic vocabulary-constrained filtering strategy that restricts model outputs to a predefined ontology vocabulary where applicable. By integrating retrieval-based contextual guidance with strict exact-match filtering, the proposed framework aims to improve both semantic accuracy and prediction consistency for the LLMs4OL Tasks A and B.

\section{Methodology}
This section presents the proposed \textit{Offline Retrieval-Augmented Ontology Learning Framework} developed for the LLMs4OL 2026 Flagship (Task A) and Reuse (Task B) challenges. The framework combines dense semantic retrieval, instruction-guided large language model (LLM) inference, and deterministic vocabulary-aware post processing to automatically construct or extend ontologies from unstructured text.

\begin{figure}[H]
    \centering
    \includegraphics[width=0.9\linewidth]{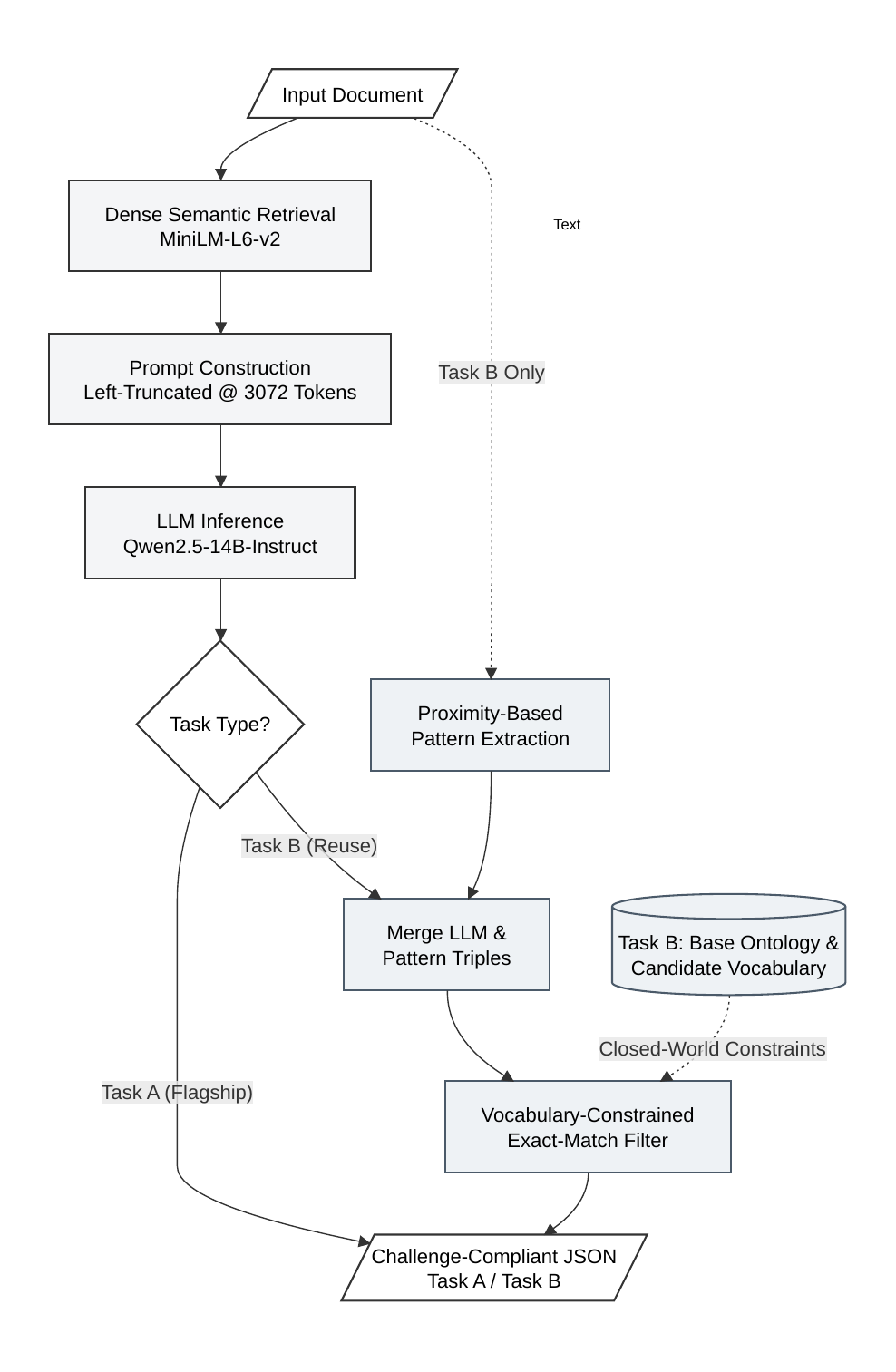}
    \caption{Architecture of the proposed Offline Retrieval-Augmented Ontology Learning Framework. The framework retrieves semantically similar examples using MiniLM embeddings, constructs retrieval-augmented prompts for Qwen2.5-14B-Instruct, and refines the generated ontology triples for Task B through vocabulary-constrained filtering before producing challenge-compliant JSON outputs.}
    \label{fig:framework}
\end{figure}

Instead of fine-tuning the language model, semantically similar training examples are dynamically retrieved from the training corpus and incorporated into the inference prompt. For the Reuse Task (Task B), the generated ontology triples are subsequently refined using a deterministic vocabulary-constrained filter before being converted into challenge-compliant JSON outputs. Figure~\ref{fig:framework} summarizes the overall workflow of the proposed framework.

\subsection{Problem Formulation}

Let

\begin{equation}
D=\{d_1,d_2,\ldots,d_n\},
\end{equation}

where $D$ denotes a collection of input documents containing unstructured natural language text.

For the \textbf{Flagship Task (Task A)}, the objective is to construct a primitive ontology represented as

\begin{equation}
O=\{(h,r,t)\},
\end{equation}

where $h$, $r$, and $t$ denote the head entity, ontology relation, and tail entity, respectively.

For the \textbf{Reuse Task (Task B)}, an existing ontology $O_{\mathrm{base}}$ is provided together with new textual evidence. The objective is to infer only the ontology triples required to extend the existing ontology, rather than reconstructing it from scratch. Formally,

\begin{equation}
O_{\mathrm{final}}
=
O_{\mathrm{base}}
\cup
O_{\mathrm{new}},
\end{equation}

where $O_{\mathrm{new}}$ denotes the newly inferred ontology triples extracted from the input document. Consequently, the proposed framework supports both ontology construction and ontology extension within a unified inference pipeline.

\subsection{Dense Semantic Retrieval}
To provide contextual guidance during inference, semantically similar ontology learning examples are retrieved from the training corpus using dense semantic representations. Each training document is encoded with the \texttt{sentence-transformers/all-MiniLM-L6-v2} model to obtain a fixed-length embedding. Let
\begin{equation}
e_{i} = f(d_{i}),
\end{equation}
where $f(\cdot)$ denotes the sentence embedding model and $e_{i}$ represents the embedding of the $i^{\text{th}}$ training document. Similarly, the embedding of the input document $q$ is computed as
\begin{equation}
e_{q} = f(q).
\end{equation}
The semantic similarity between the input document and each indexed example is computed using cosine similarity:
\begin{equation}
\text{sim}(q, d_{i}) = \frac{e_{q} \cdot e_{i}}{\|e_{q}\| \|e_{i}\|}.
\end{equation}
Training documents are ranked according to their similarity scores, and the most relevant examples are selected as in-context demonstrations for prompt construction, with the number of retrieved demonstrations tuned separately per task: top-5 for Task A and top-2 for Task B. Task A retrieves more demonstrations because the flagship setting requires the model to jointly reason across term extraction, typing, taxonomy, and relation extraction from raw text with no provided vocabulary, so broader in-context coverage was found to help; Task B provides an explicit candidate vocabulary (terms and types) and a partial ontology as additional grounding context, so fewer demonstrations were needed while also keeping prompts shorter and reducing truncation risk. Unlike static few-shot prompting, this retrieval strategy dynamically adapts the demonstrations to each input document, providing task-specific guidance without modifying the parameters of the underlying language model.

\subsection{Prompt Construction and Ontology Generation}
The retrieved examples are incorporated into a structured prompt for ontology generation using Qwen2.5-14B-Instruct. The system prompts explicitly guide the model's behavior. For Task A, the prompt instructs the model: "You are an expert ontology engineer. Extract ALL ontology triples from the passage. A triple is [subject, relation, object]... Copy each subject and object EXACTLY as it appears in the text." It also maps natural language phrases directly to a closed set of relations (e.g., "subclass of" maps to "is-a"). For Task B, the prompt enforces strict hierarchy constraints: "Predict only DIRECT parent relationships — connect a term to its MOST SPECIFIC correct type."

Generation utilizes Qwen2.5-14B-Instruct in 16-bit precision with greedy decoding (sampling disabled) for deterministic outputs, a batch size of 10, and task-specific output budgets: 900 maximum new tokens for Task A (to accommodate the larger number of triples and prevent truncation on the ~10 longest samples) and 320 maximum new tokens for Task B.

Although Qwen2.5-14B-Instruct supports a default context window of 32,768 tokens, we empirically restricted the maximum input length to 3,072 tokens. This threshold was selected to bound the Key-Value (KV) cache memory footprint and cap logits scaling spikes during batched inference on 80GB VRAM hardware. To ensure critical instructions were never lost, we employed a left-truncation strategy. If a dynamically constructed prompt exceeded 3,072 tokens, the oldest retrieved in-context demonstrations were dropped first, guaranteeing that the system instructions and the target query at the end of the prompt remained completely intact.

The generated response is parsed into ontology triples using a three-stage fallback parser: direct JSON parsing, regex extraction of the first well-formed array, and line-by-line regex recovery of quoted triples as a last resort. This improves robustness to minor formatting deviations (e.g. the model wrapping its JSON output in prose) without inferring or altering any content beyond what the model generated. Duplicate triples are removed, along with trivial reflexive self-loops.
\subsection{Hybrid Extraction and Vocabulary-Constrained Filtering (Task B)}
Task B provides each sample with an explicit closed vocabulary of candidate terms and types, together with a partial ontology. To respect this closed-world assumption and maximize performance, we implement a two-stage post-processing pipeline for Task B.First, to mitigate the inherent limitations of generative LLMs—particularly regarding omitted classifications—we supplement the LLM predictions with deterministic pattern-matching heuristics. We employ a proximity-based term-typing heuristic that scans the document for sentence-level co-occurrences of predefined terms and types, assigning an instance-of relation when the lexical distance between them falls within a high-confidence threshold. The triples extracted via this heuristic are merged with the generative LLM outputs to serve as a critical recall booster.Second, the merged triples are passed through a deterministic vocabulary-constrained filter. A candidate triple $[h, r, t]$ is retained only if at least one endpoint ($h$ or $t$) is a member of the sample’s known vocabulary—the union of the provided terms, the provided types, and all entities already appearing in the sample’s initial ontology triples. Triples that duplicate a relationship already present in the initial ontology are discarded. This is a strict exact-match membership filter; a triple whose endpoints do not exactly match a known vocabulary entry is dropped outright rather than lexically repaired.No equivalent vocabulary constraint is applied to Task A, since the flagship setting provides no closed candidate vocabulary to constrain against.

\subsection{Output Generation}
The validated ontology triples, after the generation and filtering steps, are transformed to the official JSON format required by the LLMs4OL evaluation framework. For the Flagship Task (Task A), the framework produces the generated primitive ontology triples for each input document. For the Reuse Task (Task B) only the newly inferred ontology triples are generated as an extension of the given base ontology, while keeping its original structure. The resulting json files are according to the submission schema provided by the challenge organizers, which means they can be directly evaluated without any further processing.
\section{Experiments and Results}

In this section, we show the empirical results of our proposed Retrieval-Augmented Generation (RAG) framework on End-to-End Flagship (Task A) and Ontology Extension (Task B) challenges. We report performance measured with the official challenge evaluation script which computes task-oriented metrics in parallel with Graph Similarity (including Edge F1, Neighborhood Similarity and Taxonomy Similarity). To evaluate the generated ontologies, the script calculates three variants of matches: Exact Match enforces strict string equality, requiring that the predicted head, relation, and tail lexical forms match exactly the gold standard. Fuzzy Match relaxes this lexical constraint by utilizing RapidFuzz string alignment, penalizing minor morphological variations or token mismatches based on a calculated Levenshtein distance ratio. Finally, Semantic Match relies on continuous embedding similarity, accepting predictions that are conceptually equivalent to the gold standard even if they differ entirely in surface-level string representation.

\subsection{Results on Task B: Ontology Extension (Reuse)}

The Ontology Extension task evaluates the system's ability to incrementally expand a partial ontology given a constrained set of valid terms and types. Our implementation, driven by the vocabulary-constrained filtering algorithm, achieved exceptionally high structural compliance.

\begin{table}[ht]
\centering
\caption{Evaluation Metrics for Task B (Reuse Task): Graph Similarity}
\label{tab:taskb_graph}

\begin{tabularx}{\linewidth}{|>{\raggedright\arraybackslash}p{3cm}|
>{\centering\arraybackslash}X|
>{\centering\arraybackslash}X|
>{\centering\arraybackslash}X|
>{\centering\arraybackslash}X|}
\hline
\textbf{Metric Category} &
\textbf{Edge F1} &
\textbf{Neighborhood Similarity} &
\textbf{Taxonomy Similarity} &
\textbf{Graph Similarity} \\
\hline

Exact Match &
0.8776 &
0.8500 &
0.8165 &
0.8480 \\
\hline

Fuzzy Match &
0.8945 &
0.8690 &
0.8332 &
0.8656 \\
\hline

Semantic Match &
0.8988 &
0.8727 &
0.8359 &
0.8692 \\
\hline

\end{tabularx}
\end{table}

\begin{table}[ht]
\centering
\caption{Task-Oriented Metrics for Task B (Reuse Task)}
\label{tab:taskb_task}

\begin{tabularx}{\linewidth}{|>{\raggedright\arraybackslash}p{3.5cm}|
>{\centering\arraybackslash}X|
>{\centering\arraybackslash}X|
>{\centering\arraybackslash}X|}
\hline

\textbf{Metric} &
\textbf{Precision} &
\textbf{Recall} &
\textbf{F1-Score} \\
\hline

Term-Typing &
0.9324 &
0.9079 &
0.9200 \\
\hline

Taxonomy-Discovery &
0.9369 &
0.7846 &
0.8540 \\
\hline

Non-Taxonomic-RE &
0.0000 &
0.0000 &
0.0000 \\
\hline

\end{tabularx}
\end{table}

The results in Table 1 demonstrate the critical importance of deterministic post-processing. Because the generative outputs were strictly filtered against the predefined canonical vocabulary, the delta between the Exact Match Graph Similarity (0.8480) and Semantic Match Graph Similarity (0.8692) is remarkably narrow. This confirms that the model successfully avoided out-of-vocabulary hallucinations. The framework excelled particularly in Term-Typing, achieving a robust 0.9200 F1-score, suggesting that context-augmented prompting coupled with strict vocabulary-constrained filtering contributes positively to closed-world classification. We note, however, that the absence of strict ablation studies limits our ability to precisely quantify the isolated performance gains provided by the RAG module versus the rule-based extraction and filtering components.

\subsection{Results on Task A: End-to-End Flagship Task}

The Flagship task is fundamentally more complex, requiring the model to extract and structure the primitive ontology without predefined vocabulary constraints.

\begin{table}[ht]
\centering
\caption{Evaluation Metrics for Task A (Flagship Task): Graph Similarity}
\label{tab:taska_graph}

\begin{tabularx}{\linewidth}{|>{\raggedright\arraybackslash}p{3cm}|
>{\centering\arraybackslash}X|
>{\centering\arraybackslash}X|
>{\centering\arraybackslash}X|
>{\centering\arraybackslash}X|}
\hline

\textbf{Metric Category} &
\textbf{Edge F1} &
\textbf{Neighborhood Similarity} &
\textbf{Taxonomy Similarity} &
\textbf{Graph Similarity} \\
\hline

Exact Match &
0.5891 &
0.5152 &
0.5017 &
0.5353 \\
\hline

Fuzzy Match &
0.7365 &
0.6765 &
0.6538 &
0.6889 \\
\hline

Semantic Match &
0.7854 &
0.7319 &
0.7076 &
0.7416 \\
\hline

\end{tabularx}
\end{table}

\begin{table}[h!]
\centering
\caption{Task-Oriented Metrics for Task A (Flagship Task)}
\label{tab:taska_task}

\begin{tabularx}{\linewidth}{|>{\raggedright\arraybackslash}p{3.5cm}|
>{\centering\arraybackslash}X|
>{\centering\arraybackslash}X|
>{\centering\arraybackslash}X|}
\hline

\textbf{Metric} &
\textbf{Precision} &
\textbf{Recall} &
\textbf{F1-Score} \\
\hline

Term-Typing &
0.5952 &
0.4245 &
0.4956 \\
\hline

Taxonomy-Discovery &
0.6161 &
0.5708 &
0.5926 \\
\hline

Non-Taxonomic-RE &
0.0000 &
0.0000 &
0.0000 \\
\hline

\end{tabularx}
\end{table}

In the unconstrained setting of Task A, we observe a significant divergence between exact and semantic metrics. The Exact Match Graph Similarity drops to 0.5353, while the Semantic Match Graph Similarity reaches a much stronger 0.7416. This 20-point difference highlights a known limitation of unconstrained generative models: although they successfully capture the underlying semantic concepts, as reflected by the high semantic similarity score, they often fail to reproduce the exact lexical forms required by strict exact-match evaluation protocols.

\subsection{In-Depth Analysis: The Non-Taxonomic Extraction Bottleneck}
A review of the evaluation logs across both Task A and Task B shows that the F1-score for Non-Taxonomic Relation Extraction is exactly 0.0000 in both tracks. Two factors contribute to this result.  First, our relation-mapping instructions defined a closed set of six strictly taxonomic and typing relation strings (is-a, instance-of, disjoint with, equivalent class, part of, has part). This design choice was empirical: early exploration of the training data revealed that hierarchical relations overwhelmingly dominated the dataset. To prevent the LLM from hallucinating diverse, unstandardized predicates, we restricted its generation space. By forcing the LLM to map all textual relationships to this strict set (e.g., standardizing "type of" and "category of" to "is-a"), we implicitly handled relational redundancies directly during the generation phase, which is why we did not employ a post-processing predicate filter to catch impossible co-occurring relations. However, under this strict prompt design, the model had no mechanism by which to emit a non-taxonomic relation label.  Second, the extent to which this constrains performance depends on how prevalent genuinely non-taxonomic relations are in the gold data itself for these two tasks. Without a full accounting of the gold relation-type distribution, we cannot fully separate our prompt-design limitation from a property of the benchmark data. We report the 0.0 score as an observed result and flag the closed relation vocabulary as the most direct, verifiable explanation available from our pipeline.

\section{Conclusion}
In this work, we presented an offline Retrieval-Augmented Generation (RAG) framework utilizing Qwen2.5-14B-Instruct and sentence-transformers/all-MiniLM-L6-v2 for the LLMs4OL 2026 Challenge. By combining left-side context window management with deterministic vocabulary-constrained filtering for Task B, our approach effectively prevented out-of-vocabulary hallucinations from entering the final generated outputs and stabilized overall predictions. Our system achieved strong results on Task B (Ontology Extension), securing an Overall Semantic Graph Similarity of 0.8692, a Term-Typing F1-score of 0.9200, and a Taxonomy Discovery F1-score of 0.8540, while also demonstrating the effectiveness of retrieval-augmented prompting in unconstrained scenarios by achieving an Overall Semantic Graph Similarity of 0.7416 on Task A (End-to-End Flagship). However, error analysis revealed a major bottleneck: a 0.0000 F1-score in Non-Taxonomic Relation Extraction across both tasks, highlighting both a limitation of our current prompt design---which defines only a closed taxonomic/typing relation vocabulary---and an open question about how prevalent non-taxonomic relations are in the benchmark data itself. Moving forward, future work will focus on hybrid neuro-symbolic integration, schema-guided constrained decoding, and ontology-aware relation extraction techniques to improve non-taxonomic relation prediction. Overall, the proposed framework demonstrates that combining retrieval augmentation with deterministic vocabulary alignment provides a practical and effective approach for ontology learning using large language models, while also identifying key challenges that motivate future research in hybrid ontology construction methods.

\section*{Limitations}
While the proposed framework achieved strong results, several limitations remain. First, the individual performance contributions of dynamic retrieval, prompt configuration, and hybrid rule-based extraction were not isolated through formal ablation studies, making it difficult to definitively quantify each component's distinct impact. Second, our post-processing focused entirely on entity vocabulary filtering rather than explicit semantic predicate filtering. Finally, critical hyperparameters—such as the 3,072-token truncation limit and the varying in-context demonstration counts (top-5 vs. top-2)—were set empirically based on hardware constraints and preliminary dev-set evaluations, rather than through exhaustive grid search optimization.

\section*{Data availability statement}
The dataset used in this study was provided by the task organizers as part of the LLMs4OL 2026 challenge.

\section*{Underlying and related material}
The datasets analyzed during the current study are available in the official LLMs4OL 2026 Challenge repository. The code used to generate the submitted results is available at: \url{https://github.com/shivam07041/team-pro-integration_OntoLearner_Submission}.

\section*{Funding}
This research was supported by the Startup Research Grant provided by the Department of Information Technology, Indian Institute of Information Technology Allahabad.

\section*{Author contributions}
\textbf{Shivam Mishra}: Conceptualization, Methodology, Software, Writing. \textbf{Dhannu Ram Meena}: Software, Investigation, Methodology. \textbf{Muneendra Ojha}: Supervision, Writing -- review \& editing. \textbf{Krishna Pratap Singh}: Supervision, Writing -- review \& editing. \textbf{Kuldeep Singh}: Resources, Investigation, Software.

\section*{Competing interests}
The authors declare no competing interests.
\section*{Acknowledgment}
The authors sincerely thank Dr. Sumit Singh and Arindam Ghosh for assistance during this research work. 

\section*{Declaration on Generative AI}
Generative AI tools were used to assist with manuscript drafting and language refinement. The authors reviewed and validated all AI-assisted content and take full responsibility for the accuracy, originality, and integrity of the final manuscript.

\printbibliography[heading=references]

\end{document}